\documentclass[letterpaper, 10 pt, conference]{ieeeconf}  

\IEEEoverridecommandlockouts                              

\usepackage{graphicx} 
\usepackage{amsmath} 
\usepackage{amssymb}  
\usepackage{cite}
\usepackage{hyperref} 
\usepackage{subcaption}
\usepackage{hyperref}
\usepackage{caption}
\usepackage{float}

\newcommand{\comment}[1]{}
\usepackage{color}

\title{\LARGE \bf
AquaMILR+: Design of an untethered limbless robot for complex aquatic terrain navigation
}

\author{%
Matthew Fernandez$^{1,\dagger}$, Tianyu Wang$^{1,\dagger}$, Galen Tunnicliffe$^{1}$, Donoven Dortilus$^{1}$,
Peter Gunnarson$^{2}$, \\John O. Dabiri$^{2}$, Daniel I. Goldman$^{1}$%
\thanks{$\dagger$These authors contributed equally to this work.}
\thanks{$^{1}$Matthew Fernandez, Tianyu Wang, Galen Tunnicliffe, Donoven Dortilus and Daniel I. Goldman are with Georgia Institute of Technology, Atlanta, GA 30332, USA. {\tt\small \{mfernandez64, tianyuwang, gtunnicliffe3, ddortilus3\}@gatech.edu, daniel.goldman@physics.gatech.edu}}%
\thanks{$^{2}$Peter Gunnarson and John O. Dabiri are with California Institute of Technology, Pasadena, CA 91125, USA. {\tt\small \{pjgunnar, jodabiri\}@caltech.edu}}%
}

\begin{document}
\maketitle
\thispagestyle{empty}
\pagestyle{empty}

\begin{abstract}
This paper presents AquaMILR+, an untethered limbless robot designed for agile navigation in complex aquatic environments. The robot features a bilateral actuation mechanism that models musculoskeletal actuation in many anguilliform swimming organisms which propagates a moving wave from head to tail allowing open fluid undulatory swimming. This actuation mechanism employs mechanical intelligence, enhancing the robot's maneuverability when interacting with obstacles. AquaMILR+ also includes a compact depth control system inspired by the swim bladder and lung structures of eels and sea snakes. The mechanism, driven by a syringe and telescoping leadscrew, enables depth and pitch control -- capabilities that are difficult for most anguilliform swimming robots to achieve. Additional structures, such as fins and a tail, further improve stability and propulsion efficiency. Our tests in both open water and indoor 2D and 3D heterogeneous aquatic environments highlight AquaMILR+'s capabilities and suggest a promising system for complex underwater tasks such as search and rescue and deep-sea exploration.
\end{abstract}

\section{Introduction}\label{sec:intro}

\begin{figure}[t]
\centering
\includegraphics[width=0.76\columnwidth]{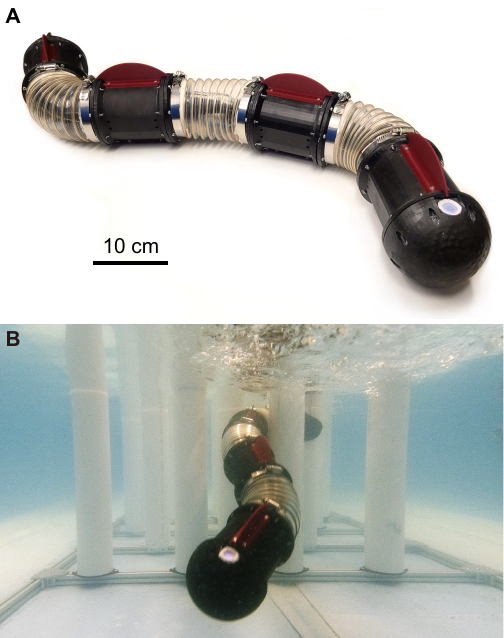}
\caption{The aquatic limbless robot AquaMILR+ designed for locomotion in complex and cluttered environments. (A) Full robot assembly, featuring a modular self-contained untethered architecture. (B) AquaMILR+ navigating a laboratory obstacle-rich environment (vertical posts).}
\label{fig:intro_robot}
\vspace{-1.5em}
\end{figure}

The development of novel underwater robots offers promising solutions for a variety of applications, including deep-sea exploration, search and rescue, infrastructure inspection, environmental monitoring, and underwater construction. Among these, autonomous underwater vehicles (AUVs) have been extensively studied, demonstrating their reliability and efficacy for basic tasks. However, their traditionally rigid design can hinder performance in confined or cluttered environments~\cite{watson2020localisation}.

Beyond traditional AUVs, various other types of underwater robots have been developed to address these challenges across different application scenarios~\cite{spino2024towards,chu2012review,qu2024recent,wang2020development,raj2016fish,bogue2015underwater}. Moreover, bio-inspired designs have significantly improved swimming speed and capability for specific tasks~\cite{wang2023versatile,thandiackal2021emergence,ren2021research,baines2022multi,katzschmann2018exploration,zhu2019tuna}. Despite advancements in bioinspired and engineered designs, many underwater robots still struggle with agility and robustness in cluttered or dynamic aquatic environments~\cite{watson2020localisation,wong2018autonomous}.

\begin{figure*}[t]
\centering
\includegraphics[width=1\textwidth]{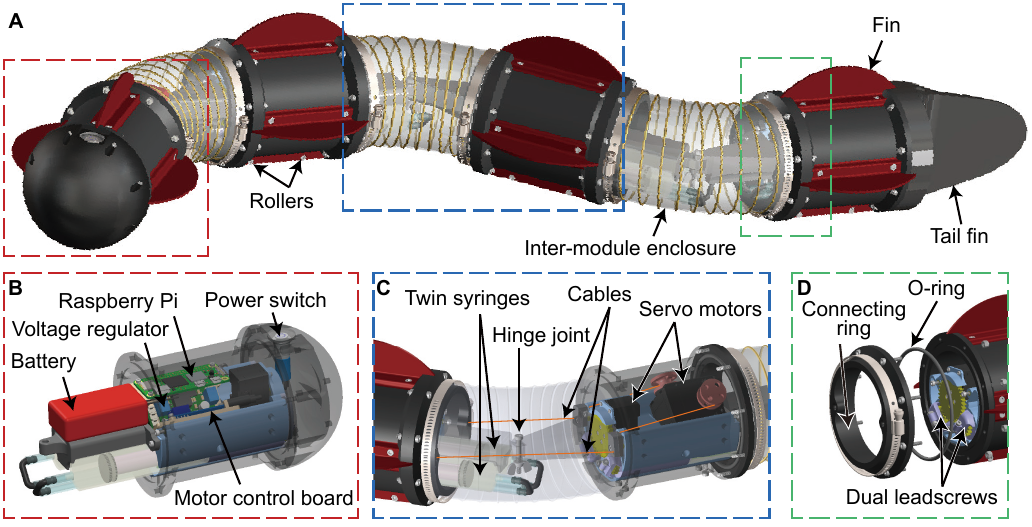}
\caption{Detailed design of AquaMILR+. (A) An assembly of 4 modules with 3 joints. (B) The electronics module contained within the head module, features onboard power, a single-board computer, and a waterproof power switch. (C) An internal diagram of each module and inter-module enclosure, including the depth control and cable-driving servo motors, cable routing, and revolute joint. (D) The primary waterproofing method between modules, including a gasket seal to clamp modules with in between O-ring.}
\label{fig:robot_design}
\vspace{-1.5em}
\end{figure*}

Limbless robots have emerged as a promising solution for navigating complex terrains, particularly in confined terrestrial spaces, thanks to their hyper-redundant body structures~\cite{wang2023mechanical,kojouharov2024anisotropic,wang2020directional}. Inspired by anguilliform swimmers such as eels and sea snakes, limbless robots have shown potential for similar applications in underwater environments~\cite{ACM_R5H_2010,yamada2005development,liljeback2014mamba,crespi2008online,li2023underwater}. However, while their many degrees of freedom offer flexibility, they can also introduce challenges--most current limbless underwater robots lack sufficient depth control and precise body shape modulation, limiting their ability to effectively navigate dynamic and obstacle-rich environments.

In this paper, we introduce AquaMILR+ (Fig.~\ref{fig:intro_robot}), an untethered, limbless underwater robot designed to overcome these challenges (built upon previous works~\cite{wang2023mechanical,wang2024aquamilr}). AquaMILR+ incorporates a bilateral actuation mechanism inspired by the musculoskeletal movements of anguilliform swimmers, enhancing its maneuverability and agility when interacting with obstacles. Additionally, the robot features a depth control system that enables dynamic vertical and pitch adjustments. Using this compact and distributed water-exchange system, AquaMILR+ demonstrates its effective buoyancy control while undulating. Our evaluations show that AquaMILR+ excels in both open water and cluttered environments, demonstrating promising ability for a wide range of applications, including search and rescue, marine exploration, and environmental monitoring.


\section{Robot design}
\label{sec:design}

\subsection{Design overview}
AquaMILR+ is an untethered cable-driven undulatory robot with full control over body shape and depth, featuring programmable body compliance (Fig.~\ref{fig:robot_design}A). It has a body length of 1 m, with 4 primary modules and 3 joints. The robot's fully waterproof design includes completely onboard power and communication enclosed in the head module, and a distributed depth control system in each module. It also holds appendages including multiple fins, a tail, and ventral rollers for effective in-water propulsion and stability.

\begin{table}[h]
\centering
\renewcommand{\arraystretch}{1.1} 
\begin{tabular}{l l} 
\hline\hline 
\\[-1.25em] 
\textbf{Mass} & \begin{tabular}[c]{@{}l@{}}Single Module: 0.68 kg \\ Full 4 Module Robot: 6.15 kg \end{tabular} \\ 
\hline
\textbf{Dimensions} & \begin{tabular}[c]{@{}l@{}}10 cm Diameter\\ 101.3 cm Length\end{tabular} \\ 
\hline
\textbf{Power} & 11.1 V, 1.0 A (normal operation) \\ 
\hline
\textbf{Communication} & \begin{tabular}[c]{@{}l@{}}Wi-Fi (external)\\ RS-485 Serial (internal)\end{tabular} \\ 
\hline
\textbf{Depth} & $>$1.52 m (maximum depth tested) \\ 
\hline
\textbf{Sensing} & \begin{tabular}[c]{@{}l@{}}Individual cable length and tension\\ Leadscrew position\end{tabular} \\ 
\hline
\textbf{Actuation} & \begin{tabular}[c]{@{}l@{}}Cable Motors: 1.4 Nm (stall)\\ Leadscrew Motors: 0.92 Nm (stall)\end{tabular} \\ 
\hline\hline 
\\[-1.25em] 
\end{tabular}
\caption{AquaMILR+ specifications and ratings.}
\end{table}

\subsection{Module components} 
Each of the primary resin-printed modules is 100 cm in length and 10.8 cm in diameter. Inside each module lays a PLA insert which serves as the attachment point to all components in the module assembly. This insert is hollowed out, filled with 2-mm-diameter lead balls, and sealed, creating a form-fitting weight to achieve neutral buoyancy. Additionally, the insert has bottom-side cutouts for twin syringe-leadscrew assemblies for depth control. On the top-side, it has mounts for the cable-driving servo motor and the depth-control servo motor (Fig.~\ref{fig:robot_design}C).

On the outside of the modules, each side has four mounting holes in the resin holding screw-to-expand threaded inserts allowing for easy fastening of fins on all sides. On the bottom side, a set of four low-profile acrylic rollers are attached to provide anisotropic ventral friction allowing AquaMILR+ to effectively move on the seabed and hard ground~\cite{rieser2019dynamics}. The pectoral fins on the sides are helpful for robot stability and pitch control in relatively open-water areas but can be removed to allow dense obstacle navigation.

\subsection{Waterproofing}
AquaMILR+ features a robust, waterproof design which allows for safe operation at depth. External module casings and mating parts were made from resin on the Elegoo Saturn 3 Ultra MSLA 3D printer, preventing the possibility of water seepage through layer lines. The primary sealing method is a flange-gasket system, where each side of the module casings has a gland to hold an 8.9-cm-diameter O-ring (Fig.~\ref{fig:robot_design}D). The connecting ring includes an 8-screw circular pattern which clamps against the O-ring, providing even pressure.

These connecting rings are a part of the inter-module enclosure, which includes an 8.9-cm-diameter PVC flex-duct ventilation tubing fixed to the resin connecting ring on each side for a total length of 15 cm  (Fig.~\ref{fig:robot_design}A). These parts are fixed and sealed together using 3M Marine Adhesive Sealant 5200 at their intersection. On top of these seals, an 8.9-cm stainless steel hose clamp is fastened as a mechanical constraint taking any potential load off of the sealant. Flex-duct tubing was selected because of its extensibility, essential to allow the collapse of one side and the extension of another as the joint bends. Its spring steel reinforcement maintains the structure of this flexible membrane, resulting in a relatively constant volume through different depths to not dramatically influence buoyancy. This inter-module enclosure offers a simple watertight solution for wiring between modules as well as joint actuation solution does add volume (and thus requires extra mass for neutral buoyancy) to the system, the extra space allows for the depth control system, weight storage, and a reliable static seal.

\subsection{Bilateral cable actuation}
To enable locomotion, AquaMILR+ includes a bilateral actuation mechanism, leveraging a dual pulley-cable system at each joint (Fig.~\ref{fig:robot_design}C). Each module contains a Robotis Dynamixel 2XL430-W250-T dual-axis servo motor with a 10-mm-diameter pulley connected to each axis. In between each module is a single-degree-of-freedom revolute joint, where a male and female PLA component are fastened to their respective module inserts and held together by an M4 shoulder bolt with wingnut. Additionally, this joint assembly features detachable PLA cable guides to relieve tension on the resin components. Rikimaru braided fishing line (800 N strength) is tied to each pulley and routed through their respective joint's cable guide and connecting rings, ensuring the calculated cable path. The cable is finally tied to the opposite connecting ring on the next module allowing the two cables to control the angle of the revolute joint between each module.

\subsection{Power and communication}
Different from the other modules, the head module contains all of the power and communication components 
(Fig.~\ref{fig:robot_design}B). When combining this with the RS-485 serial communication protocol used by all of the Dynamixel servo motors, AquaMILR+ maintains 3 wires running along its body length with a simple disconnect at each joint. This structure enforces the modularity of the design, where to extend the robot length or disassemble it for servicing, modules can be simply connected in series to one another.

On this electronics tray, there are four main boards tightly packed in the head with an 11.1 V 1000 mAh LiPo battery fixed to the first joint. This battery provides its full voltage to the servo motors through the Robotis power distribution board and then splits off to a voltage regulator that outputs 5 V, protecting the rest of the electronics. From here, 5 V is provided to a Raspberry Pi Zero 2W which serves as the main computer for AquaMILR+, containing the relevant libraries and sequences. The Raspberry Pi is then connected to the Robotis U2D2 motor control board, which handles the bulk motor reading and writing commands with RS-485 serial communication protocol. With this power setup and consumption of around 1.0 A under normal operation, AquaMILR+ can run continuously for one hour before requiring battery replacement. External to AquaMILR+, the Raspberry Pi connects through Wi-Fi to an external computer so that the operator can remotely control the untethered robot via remote desktop (RealVNC)
Finally, the battery is wired in series with the main power switch mounted on the outside top surface of the rounded head attachment (Fig.~\ref{fig:robot_design}B). This switch also includes a visible LED allowing visual confirmation of activation and battery life. On top of this switch, there is a silicone cover cast from Smooth-On ECOFLEX 00-35 Fast. This is sealed over the switch with J-B Weld Marine Epoxy allowing the switch to be externally activated without compromising AquaMILR+'s seals.

\begin{figure}[t]
\centering
\includegraphics[width=1\columnwidth]{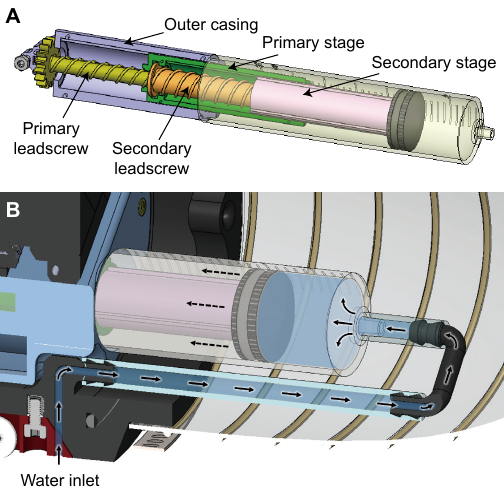}
\caption{Self-contained depth control system. (A) A telescopic leadscrew design for syringe activation, granting extra stroke in a compact space. (B) The water channel used by the syringes to change AquaMILR+'s buoyancy.}
\label{fig:variableBuoyancyfig}
\vspace{-1.5em}
\end{figure}

\subsection{Depth control system}
Through the use of local fluid exchange with surrounding waters~\cite{katzschmann2018exploration,tiwari2020design}, AquaMILR+ features a self-contained depth control system that maintains a constant external volume. This is accomplished through the use of a dual-syringe water exchange system in every module which connects to open water just underneath the modules (Fig.~\ref{fig:variableBuoyancyfig}B). The robot mass was carefully calibrated to achieve neutral buoyancy, allowing a small volume of fluid displacement to impact the robot's acceleration. This calibration was achieved by adding sealed packets of 2-mm-diameter lead balls in each module and inter-module enclosure, allowing easy adjustments based on environment or attachment changes. Syringes were chosen due to their preferred form factor, simple actuation, and limited inertial consequences as a result of their inextensibility. These syringes each have a 35-mL internal volume with a 24-mm diameter, maximizing the available module space.

Despite pre-existing architectures on syringe actuation~\cite{spino2024towards}, traditional linear actuation systems struggle to remain volume efficient, where the primary actuator is often longer than the stroke length needed. To avoid this and allow a greater maximum depth acceleration of AquaMILR+, a novel FDM printed telescopic leadscrew mechanism was developed (Fig.~\ref{fig:variableBuoyancyfig}A). This leadscrew mechanism operates similarly to traditional leadscrews but with a cascading secondary stage allowing a 60\% increase in stroke length in comparison. In this design, the primary leadscrew (yellow) has a slotted key with the secondary leadscrew (orange), effectively making them rotationally coupled while allowing motion linearly. From here, similarly to standard leadscrews, the primary leadscrew is threaded into the primary stage (green). This primary stage has 3 tabs along its circumference that mate with slots on the outer casing (lavender), which constrains the primary stage to move linearly upon rotation of the primary leadscrew. The secondary leadscrew is then able to travel linearly with the primary stage, and a secondary stage (pink) is constrained similarly as it is threaded with the secondary leadscrew and slotted with the primary stage. When combined, the driving of the primary leadscrew results in the telescoping of the mechanism's two stages for a greater overall stroke (see \href{https://youtu.be/l1MjoG7HlX8}{supplementary movie}). 

To adapt to the syringe, the outer casing is press-fit into the syringe's inner diameter and the secondary stage is adapted to hold the silicone plunger (black) from the syringe. The primary leadscrew is axially constrained with the outer casing to keep the mechanism stable, and two of these leadscrew assemblies are slotted and locked into each module insert. Finally, the two leadscrew assemblies in each module are driven by a Robotis Dynamixel XC330-T288-T servo motor attached to a 36-tooth spur gear (Fig.~\ref{fig:robot_design}C). This spur gear meshes with the 16-tooth spur gear at the end of each primary leadscrew, controlling both syringes.

\section{Robot control}
\subsection{Bilateral body actuation}
\subsubsection{Suggested gait}

To implement a basic traveling-wave locomotion pattern in AquaMILR+, we employed a shape control scheme following the ``serpenoid" curve template, originally introduced by~\cite{hirose1993biologically}. This approach enables a wave to propagate from the robot's head to its tail, governed by the following equation for the $i$-th joint angle, $\alpha_i$, at time $t$:
\begin{equation}
\alpha_i(t) = A\sin\left(2\pi\xi \frac{i}{N} - 2\pi\omega t\right) + \varphi,
\label{eq:template}
\end{equation}
where $A$ represents the amplitude, $\xi$ is the spatial frequency, $\omega$ is the temporal frequency, $\varphi$ is the angle offset, $i$ denotes the joint index, and $N$ is the total number of joints. This value of $\alpha_i$ is referred to as the ``suggested" joint angle.

\begin{figure}[t]
\centering
\includegraphics[width=0.8\columnwidth]{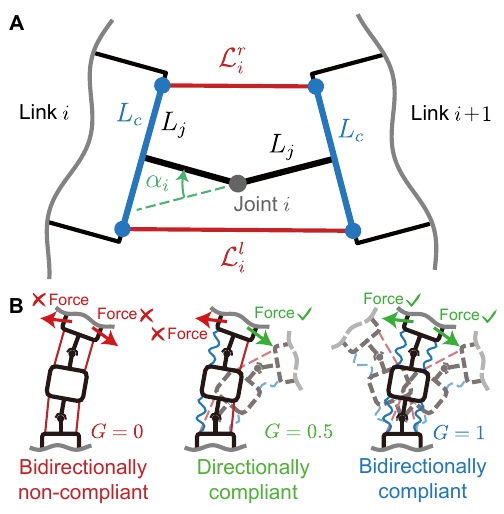}
\caption{Programmable body compliance through bilateral cable actuation mechanism. (A) A geometric model illustrating a single joint, used to determine the exact lengths of the left and right cables necessary to deploy a specified joint angle. (B) A schematic displaying various compliance states based on the generalized compliance variable $G$. Figures adapted from~\cite{wang2023mechanical}.}
\vspace{-1.5em}
\label{fig:explain_G}
\end{figure}

To accurately achieve the joint angle $\alpha$ as defined in Eq.~\ref{eq:template}, the lengths of the left and right cables around each joint must be properly adjusted, ensuring appropriate shortening of both sides (as depicted in Fig.~\ref{fig:explain_G}A). Since we utilize nonelastic cables, their deformation is negligible. The lengths of the left cable ($\mathcal{L}^l$) and right cable ($\mathcal{L}^r$) are dictated by the robot's geometry, and they follow these expressions:
\begin{equation}
\begin{aligned}
    \mathcal{L}^l(\alpha) &= 2\sqrt{L_{c}^2 + L_{j}^2} \cos\left[-\frac{\alpha}{2}+\tan^{-1}\left(\frac{L_{c}}{L_{j}}\right)\right],\\
    \mathcal{L}^r(\alpha) &= 2\sqrt{L_{c}^2 + L_{j}^2} \cos\left[\frac{\alpha}{2}+\tan^{-1}\left(\frac{L_{c}}{L_{j}}\right)\right].
\end{aligned}
\label{eq:ExactLength}
\end{equation}

\subsubsection{Programmable body compliance}
Using Eq.~\ref{eq:ExactLength}, we can precisely implement body postures for lateral undulation gaits in AquaMILR+. The bilateral actuation mechanism offers the additional benefit of programming body compliance by selectively loosening the cables. By applying the generalized compliance variable ($G$), as defined in~\cite{wang2023mechanical}, we can quantify the robot's body compliance. The cable length control scheme adjusts the lengths of each pair of left and right cables ($L_i^l$ and $L_i^r$) according to the corresponding suggested angle ($\alpha_i$):
\begin{equation}
\begin{array}{l}
L_{i}^l(\alpha_{i}) = \left\{\begin{array}{llc}{\mathcal{L}_{i}^l(\alpha_{i}),} & {\text{if } \alpha_{i} \leq -\gamma} \\ {\mathcal{L}_{i}^l[-\min(A, \gamma)]+l_0\cdot[\gamma + \alpha_{i}],} & {\text{if } \alpha_{i} > -\gamma}\end{array}\right. \\ 
L_{i}^r(\alpha_{i}) = \left\{\begin{array}{llc}{\mathcal{L}_{i}^r(\alpha_{i}),} & {\text{if } \alpha_{i} \geq \gamma} \\ 
 {\mathcal{L}_{i}^r[\min(A, \gamma)]+l_0\cdot[\gamma - \alpha_{i}],} & {\text{if } \alpha_{i} < \gamma}\end{array}\right.
\end{array}
\label{eq:policy}
\end{equation}
Here, the superscripts $l$ and $r$ refer to the left and right cables, and $\gamma = (2G_i - 1)A$. $l_0$ is a design parameter to tune the amount of loosening length of the cables (for more detail on deriving and selecting $l_0$ refer to~\cite{wang2023mechanical}). According to Eq.~\ref{eq:policy}, AquaMILR+ can achieve three representative compliance states by varying $G$ (Fig.~\ref{fig:explain_G}B): 1) bidirectionally non-compliant ($G=0$), where each joint strictly adheres to the suggested gait template from Eq.~\ref{eq:template}; 2) directionally compliant ($G=0.5$), where joints are allowed to deviate to form a larger angle than suggested; and 3) bidirectionally compliant ($G=1$), where joints can deviate in both directions, regulated by Eq.~\ref{eq:policy}.

\subsection{Depth and pitch control}
As displayed in Fig.~\ref{fig:robot_design}C and D, each of AquaMILR+'s modules is equipped with a dual leadscrew system attached to twin syringes which allows for swift and precise control of the robot’s overall depth and pitch, independent from the main body undulation. AquaMILR+'s net mass was carefully calibrated to be neutrally buoyant when the syringes hold 50\% of their stroke length, allowing equal capability in diving and ascension as the syringes exchange water. Throughout this exchange, the robot volume is assumed constant with a changing mass. This leads to a change of weight in each module, resulting in individual forces (weight or buoyancy) in the respective directions. When the total mass is greater than the neutrally buoyant mass, the robot will accelerate downward and vice versa. This resulting motion is opposed by drag as the robot dives through the water column. 

Additionally, AquaMILR+ has the ability to pitch its body due to having per-module mass control. By altering the syringe actuation percentages, the center of mass can move along the major axis of the robot, creating a moment with the buoyant force. This moment rotates the robot until the center of mass is positioned below the centroid. These relationships allow for both general acceleration of the robot as well as fine-tuned pitch control due to the precision of the servo motors that actuate the syringes.

\section{Evaluation}\label{sec:results}

\subsection{Open water locomotion}

\begin{figure}[t]
\centering
\includegraphics[width=1\columnwidth]{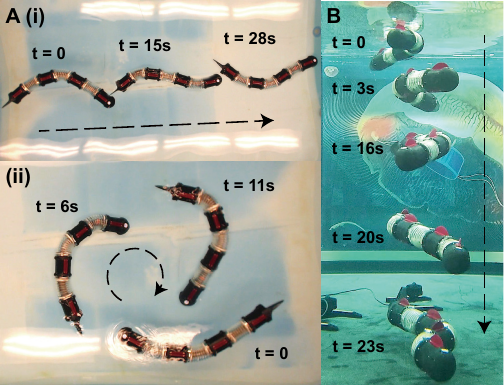}
\caption{Demonstration of locomotion and depth control capabilities of AquaMILR+. (A)(i) Straight locomotion across a 3-m-long pool; (ii) implementation of a turning gait, where the robot can turn in place with a tight sweeping area. (B) A demonstration of a controlled, slow descent to 1.52 m deep while locomoting forward.}
\label{fig:locomotion}
\vspace{-1.5em}
\end{figure}

We first evaluated AquaMILR+'s performance in an open water environment with a 3 m $\times$ 2 m $\times$ 0.5 m ($L\times W\times H$) indoor pool. In open water tests, we kept gait parameters consistent with $A = 30^\circ$, $\xi = 0.5$, $\omega = 0.2$, and $\varphi=0$ in Eq.~\ref{eq:template}, and we set $G = 0$ (noncompliant). We found that AquaMILR+ is capable of generating propulsion through this gait (Fig.~\ref{fig:locomotion}A), where the robot maintained a straight trajectory moving at $0.062 \pm 0.006$ m/s (mean $\pm$ standard deviation) and $0.305 \pm 0.031$ BL/cycle.

\begin{figure*}[t]
\centering
\includegraphics[width=0.9\textwidth]{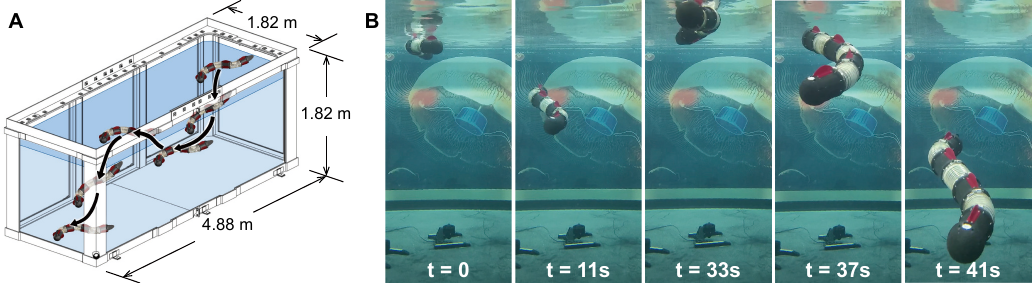}
\caption{Evaluation of AquaMILR+ locomotor capabilities, with independent depth control during undulation. (A) The path of the robot in the tank, showing control authority over movement direction. (B) Video frames throughout the locomotion from a front-camera view.}
\vspace{-1.5em}
\label{fig:depth}
\end{figure*}

To achieve turning behavior (Fig.~\ref{fig:locomotion}A), we set the offset $\varphi=20^\circ$ in Eq.~\ref{eq:template} to enable the robot to do a right offset turn~\cite{wang2020omega} where the remaining gait parameters unchanged as in the straight undulation evaluations. AquaMILR+ can turn in place with 6.21$^\circ$/s (32.1$^\circ$/cycle) within a sweeping area of an estimated radius of 0.6 m. Note that this in-place turning performance can be further improved by either increasing the offset or employing more efficient gaits such as omega turns~\cite{wang2022generalized}.

\subsection{Depth control experiments}
The depth control system of AquaMILR+ is independent of the body actuation system, allowing us to tune the robot's depth and body pitch without the concern of gait interference. We first evaluated the depth control system alone within a 4.88 m $\times$ 1.82 m $\times$ 1.82 m ($L\times W\times H$) water tank. During the evaluation, the syringes were slowly controlled from empty to full of water over 20 seconds (Fig.~\ref{fig:locomotion}B) in which the robot was able to reach a controlled descent to a depth of 1.52 m. This was done while the robot was locomoting forward, as this continuous change in volume during the gait proves essential to consider, altering the buoyancy characteristics compared to the straightened position. At this depth, the robot was still able to quickly ascend again with no compromise in seal integrity. Note that the 1.52 m depth is the maximum depth we could test with the facility; the full capability of AquaMILR+ is yet to be determined. This demonstrated the effectiveness of the depth-control system independent of the locomotion task at hand.

\begin{figure}[t]
\centering
\includegraphics[width=.9\columnwidth]{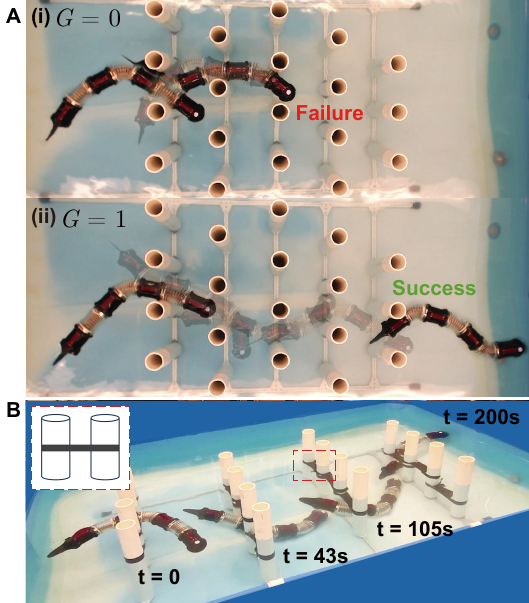}
\caption{Locomotion capability test in 2D and 3D obstacle-rich environments. (A) AquaMILR+ locomoting in a hexagonal obstacle lattice (i) without body compliance $(G=0)$, where the robot cannot traverse obstacles; and (ii) with bidirectional body compliance $(G=1)$ where the robot navigates the lattice successfully. (B) AquaMILR+ successfully navigates a complex environments with both vertical and horizontal obstacles with body compliance $(G=1)$ and buoyancy control.}
\label{fig:obstacle_2d}
\vspace{-1.5em}
\end{figure}

Understanding this, we then evaluated the robot's ability to follow more complex 3D trajectories to verify general open water performance applicable to navigation in cluttered environments. To test this, a sequence of syringe positions was commanded during the instructed undulation gait, resulting in a forward motion with changing depth to follow a complex path (Fig.~\ref{fig:depth}). With these tests, AquaMILR+ demonstrated effective locomotion in 3D aquatic environments through independent depth and gait control.

\subsection{Obstacle experiments}

To evaluate AquaMILR+'s capability of locomoting in cluttered aquatic environments, we set up an adjustable lattice structure in the indoor pool mentioned previously. To properly mount obstacles to the pool and avoid corrosion, an aluminum frame was assembled with adjustable spacing. Next, 7.6-cm-diameter PVC tubes were press-fit into PLA adapters that slide and fasten onto the aluminum frame. In this set of tests, the lattice was set up with a hexagonal pattern with 25 cm spacing.

To verify that body compliance enabled by the bilateral actuation mechanism ($G$) facilitates effective obstacle navigation, we compared the performance of the robot with no compliance ($G=0$), directional compliance ($G=0.5$), and bidirectional compliance ($G=1$). In experiments, the robot was placed in identical initial positions, and the robot was allowed to navigate the lattice until either it made it through (marked as a success), or it completely halted forward progress/overloaded the motors (marked as a failure). The gait parameters were set to $A = 50^\circ$, $\xi = 0.6$, and $\omega = 0.1$ throughout all trials. We observed that with compliance (both $G = 0.5$ and $G=1$), the robot could navigate through the lattice whereas without compliance the robot becomes stuck between obstacles frequently and failed to traverse (Fig.~\ref{fig:obstacle_2d}A), demonstrating that mechanical intelligence facilitated by bilateral actuation can improve the robot's capability of navigating obstacles without the need of sensor or motor feedback controls~\cite{wang2023mechanical,wang2024aquamilr}.

Further, we continued to evaluate AquaMILR+'s obstacle navigation now in complex, 3D environments. Similar rows of PVC on the aluminum frame were used, but with greater row spacing and added lateral obstacles, forcing a combination of locomotion with depth control to escape. During these tests, bidirectional compliance ($G=1$) was applied, along with the same gait parameters used in previous 2D lattice experiments. Throughout the test, depth had to be modulated such that the robot would avoid the lateral obstacles while staying off the pool bottom to avoid the aluminum framing. These trials successfully demonstrated AquaMILR+'s ability to navigate 3D obstacles through its independent depth control and body compliance (Fig.~\ref{fig:obstacle_2d}B). 

Video clips of robot experiments can be found in the \href{https://youtu.be/l1MjoG7HlX8}{supplementary movie}.

\vspace{-0.5em}
\section{Conclusion}\label{sec:conclusion}
In this work, we present the development of AquaMILR+, an aquatic limbless robot designed for navigating cluttered, obstacle-dense environments. The system has demonstrated watertight integrity at tested depths, along with precise control over body shape, depth, and locomotion direction. Additionally, AquaMILR+ highlights the advantages of mechanical intelligence through its body compliance, enhancing its ability to navigate obstacles. The decentralized compliance strategy employed in this platform opens the door to future advancements, such as closed-loop gait selection and tactile sensing of the surrounding environment.

Moving forward, we aim to further test AquaMILR+'s capabilities in navigating cluttered environments, focusing on both rigid and compliant obstacles in 2D and 3D settings. In parallel, we are developing a depth controller that utilizes surrounding water pressure and attitude sensing to inform the robot's current state relative to its target depth. We will also test AquaMILR+ in deeper environments until a significant decline in performance or seal integrity is observed. This work showcases an effective and scalable architecture, with potential applications in aquatic and other extreme environments.


\section{Acknowledgement}\label{sec:Acknowledgment}
The authors would like to thank Patricia Meza for assisting with graphic design and figures. This study is supported by Army Research Office grant (W911NF-11-1-0514) and National Science Foundation Physics of Living Systems Student Research Network (GR10003305). Competing interest: a provisional patent application was filed for this work on September 17, 2024.


\bibliographystyle{IEEEtran}

\bibliography{ICRA2025_UnderwaterLimbless}

\begin{thebibliography}{10}
\providecommand{\url}[1]{#1}
\csname url@rmstyle\endcsname
\providecommand{\newblock}{\relax}
\providecommand{\bibinfo}[2]{#2}
\providecommand\BIBentrySTDinterwordspacing{\spaceskip=0pt\relax}
\providecommand\BIBentryALTinterwordstretchfactor{4}
\providecommand\BIBentryALTinterwordspacing{\spaceskip=\fontdimen2\font plus
\BIBentryALTinterwordstretchfactor\fontdimen3\font minus \fontdimen4\font\relax}
\providecommand\BIBforeignlanguage[2]{{%
\expandafter\ifx\csname l@#1\endcsname\relax
\typeout{** WARNING: IEEEtran.bst: No hyphenation pattern has been}%
\typeout{** loaded for the language `#1'. Using the pattern for}%
\typeout{** the default language instead.}%
\else
\language=\csname l@#1\endcsname
\fi
#2}}

\bibitem{watson2020localisation}
S.~Watson, D.~A. Duecker, and K.~Groves, ``Localisation of unmanned underwater vehicles (uuvs) in complex and confined environments: A review,'' \emph{Sensors}, vol.~20, no.~21, p. 6203, 2020.

\bibitem{spino2024towards}
P.~Spino and D.~Rus, ``Towards centimeter-scale underwater mobile robots: An architecture for capable $\mu$auvs,'' in \emph{2024 IEEE International Conference on Robotics and Automation (ICRA)}.\hskip 1em plus 0.5em minus 0.4em\relax IEEE, 2024, pp. 1484--1490.

\bibitem{chu2012review}
W.-S. Chu, K.-T. Lee, S.-H. Song, M.-W. Han, J.-Y. Lee, H.-S. Kim, M.-S. Kim, Y.-J. Park, K.-J. Cho, and S.-H. Ahn, ``Review of biomimetic underwater robots using smart actuators,'' \emph{International journal of precision engineering and manufacturing}, vol.~13, pp. 1281--1292, 2012.

\bibitem{qu2024recent}
J.~Qu, Y.~Xu, Z.~Li, Z.~Yu, B.~Mao, Y.~Wang, Z.~Wang, Q.~Fan, X.~Qian, M.~Zhang, \emph{et~al.}, ``Recent advances on underwater soft robots,'' \emph{Advanced Intelligent Systems}, vol.~6, no.~2, p. 2300299, 2024.

\bibitem{wang2020development}
R.~Wang, S.~Wang, Y.~Wang, L.~Cheng, and M.~Tan, ``Development and motion control of biomimetic underwater robots: A survey,'' \emph{IEEE Transactions on Systems, Man, and Cybernetics: Systems}, vol.~52, no.~2, pp. 833--844, 2020.

\bibitem{raj2016fish}
A.~Raj and A.~Thakur, ``Fish-inspired robots: design, sensing, actuation, and autonomy—a review of research,'' \emph{Bioinspiration \& biomimetics}, vol.~11, no.~3, p. 031001, 2016.

\bibitem{bogue2015underwater}
R.~Bogue, ``Underwater robots: a review of technologies and applications,'' \emph{Industrial Robot: An International Journal}, vol.~42, no.~3, pp. 186--191, 2015.

\bibitem{wang2023versatile}
T.~Wang, H.-J. Joo, S.~Song, W.~Hu, C.~Keplinger, and M.~Sitti, ``A versatile jellyfish-like robotic platform for effective underwater propulsion and manipulation,'' \emph{Science Advances}, vol.~9, no.~15, p. eadg0292, 2023.

\bibitem{thandiackal2021emergence}
R.~Thandiackal, K.~Melo, L.~Paez, J.~Herault, T.~Kano, K.~Akiyama, F.~Boyer, D.~Ryczko, A.~Ishiguro, and A.~J. Ijspeert, ``Emergence of robust self-organized undulatory swimming based on local hydrodynamic force sensing,'' \emph{Science robotics}, vol.~6, no.~57, p. eabf6354, 2021.

\bibitem{ren2021research}
K.~Ren and J.~Yu, ``Research status of bionic amphibious robots: A review,'' \emph{Ocean Engineering}, vol. 227, p. 108862, 2021.

\bibitem{baines2022multi}
R.~Baines, S.~K. Patiballa, J.~Booth, L.~Ramirez, T.~Sipple, A.~Garcia, F.~Fish, and R.~Kramer-Bottiglio, ``Multi-environment robotic transitions through adaptive morphogenesis,'' \emph{Nature}, vol. 610, no. 7931, pp. 283--289, 2022.

\bibitem{katzschmann2018exploration}
R.~K. Katzschmann, J.~DelPreto, R.~MacCurdy, and D.~Rus, ``Exploration of underwater life with an acoustically controlled soft robotic fish,'' \emph{Science Robotics}, vol.~3, no.~16, p. eaar3449, 2018.

\bibitem{zhu2019tuna}
J.~Zhu, C.~White, D.~K. Wainwright, V.~Di~Santo, G.~V. Lauder, and H.~Bart-Smith, ``Tuna robotics: A high-frequency experimental platform exploring the performance space of swimming fishes,'' \emph{Science Robotics}, vol.~4, no.~34, p. eaax4615, 2019.

\bibitem{wong2018autonomous}
C.~Wong, E.~Yang, X.-T. Yan, and D.~Gu, ``Autonomous robots for harsh environments: a holistic overview of current solutions and ongoing challenges,'' \emph{Systems Science \& Control Engineering}, vol.~6, no.~1, pp. 213--219, 2018.

\bibitem{wang2023mechanical}
T.~Wang, C.~Pierce, V.~Kojouharov, B.~Chong, K.~Diaz, H.~Lu, and D.~I. Goldman, ``Mechanical intelligence simplifies control in terrestrial limbless locomotion,'' \emph{Science Robotics}, vol.~8, no.~85, p. eadi2243, 2023.

\bibitem{kojouharov2024anisotropic}
V.~Kojouharov, T.~Wang, M.~Fernandez, J.~Maeng, and D.~I. Goldman, ``Anisotropic body compliance facilitates robotic sidewinding in complex environments,'' in \emph{2024 International Conference on Robotics and Automation (ICRA)}.\hskip 1em plus 0.5em minus 0.4em\relax IEEE, 2024.

\bibitem{wang2020directional}
T.~Wang, J.~Whitman, M.~Travers, and H.~Choset, ``Directional compliance in obstacle-aided navigation for snake robots,'' in \emph{2020 American Control Conference (ACC)}.\hskip 1em plus 0.5em minus 0.4em\relax IEEE, 2020, pp. 2458--2463.

\bibitem{ACM_R5H_2010}
{Tokyo Institute of Technology} and {HiBot Corporation}, ``{ACM-R5H}: Amphibious snake robot,'' \url{https://robotsguide.com/robots/acm}, 2010.

\bibitem{yamada2005development}
H.~Yamada, S.~Chigisaki, M.~Mori, K.~Takita, K.~Ogami, and S.~Hirose, ``Development of amphibious snake-like robot acm-r5, isr2005,'' \emph{Proceedings of ISR}, p. 133, 2005.

\bibitem{liljeback2014mamba}
P.~Liljeb{\"a}ck, {\O}.~Stavdahl, K.~Y. Pettersen, and J.~T. Gravdahl, ``Mamba-a waterproof snake robot with tactile sensing,'' in \emph{2014 IEEE/RSJ International Conference on Intelligent Robots and Systems}.\hskip 1em plus 0.5em minus 0.4em\relax IEEE, 2014, pp. 294--301.

\bibitem{crespi2008online}
A.~Crespi and A.~J. Ijspeert, ``Online optimization of swimming and crawling in an amphibious snake robot,'' \emph{IEEE Transactions on robotics}, vol.~24, no.~1, pp. 75--87, 2008.

\bibitem{li2023underwater}
G.~Li, G.~Liu, D.~Leng, X.~Fang, G.~Li, and W.~Wang, ``Underwater undulating propulsion biomimetic robots: A review,'' \emph{Biomimetics}, vol.~8, no.~3, p. 318, 2023.

\bibitem{wang2024aquamilr}
T.~Wang, N.~Mankame, M.~Fernandez, V.~Kojouharov, and D.~I. Goldman, ``Aquamilr: Mechanical intelligence simplifies control of undulatory robots in cluttered fluid environments,'' \emph{arXiv preprint arXiv:2407.01733}, 2024.

\bibitem{rieser2019dynamics}
J.~M. Rieser, P.~E. Schiebel, A.~Pazouki, F.~Qian, Z.~Goddard, K.~Wiesenfeld, A.~Zangwill, D.~Negrut, and D.~I. Goldman, ``Dynamics of scattering in undulatory active collisions,'' \emph{Physical Review E}, vol.~99, no.~2, p. 022606, 2019.

\bibitem{tiwari2020design}
B.~K. Tiwari and R.~Sharma, ``Design and analysis of a variable buoyancy system for efficient hovering control of underwater vehicles with state feedback controller,'' \emph{Journal of Marine Science and Engineering}, vol.~8, no.~4, p. 263, 2020.

\bibitem{hirose1993biologically}
S.~Hirose, ``Biologically inspired robots,'' \emph{Snake-Like Locomotors and Manipulators}, 1993.

\bibitem{wang2020omega}
T.~Wang, B.~Chong, K.~Diaz, J.~Whitman, H.~Lu, M.~Travers, D.~I. Goldman, and H.~Choset, ``The omega turn: A biologically-inspired turning strategy for elongated limbless robots,'' in \emph{2020 IEEE/RSJ International Conference on Intelligent Robots and Systems (IROS)}.\hskip 1em plus 0.5em minus 0.4em\relax IEEE, 2020, pp. 7766--7771.

\bibitem{wang2022generalized}
T.~Wang, B.~Chong, Y.~Deng, R.~Fu, H.~Choset, and D.~I. Goldman, ``Generalized omega turn gait enables agile limbless robot turning in complex environments,'' in \emph{2022 International Conference on Robotics and Automation (ICRA)}.\hskip 1em plus 0.5em minus 0.4em\relax IEEE, 2022, pp. 01--07.

\end{thebibliography}

\end{document}